\documentclass{article}

\usepackage{PRIMEarxiv}

\usepackage[utf8]{inputenc} 
\usepackage[T1]{fontenc}    
\usepackage{hyperref}       
\usepackage{url}            
\usepackage{booktabs}       
\usepackage{amsfonts}       
\usepackage{amsmath}
\usepackage{nicefrac}       
\usepackage{microtype}      
\usepackage{lipsum}
\usepackage{graphicx}
\usepackage{subfigure}
\usepackage{makecell}
\usepackage{multirow}
\graphicspath{{media/}}     

\title{Improved Aggregating and Accelerating Training Methods for Spatial Graph Neural Networks on Fraud Detection
}

\author{
  Yufan Zeng \\
  College of Telecommunications \& Information Engineering \\
  Nanjing University of Posts and Telecommunications \\
  Nanjing\\
  \texttt{yufanzeng@outlook.com} \\
  \And
  Jiashan Tang* \\
  College of Science \\
  Nanjing University of Posts and Telecommunications; \\
  School of Mathematical Sciences \\
  Nanjing Normal University \\
  Nanjing\\
  \texttt{tangjs@njupt.edu.cn} \\
}

\begin{document}
\maketitle

\begin{abstract}
Graph neural networks (GNNs) have been widely applied to numerous fields. A recent work which combines layered structure and residual connection proposes an improved deep architecture to extend CAmouflage-REsistant GNN (CARE-GNN) to deep models named as Residual Layered CARE-GNN (RLC-GNN), which forms self-correcting and incremental learning mechanism, and achieves significant performance improvements on fraud detection task. However, we spot three issues of RLC-GNN, which are the usage of neighboring information reaching limitation, the training difficulty which is inherent problem to deep models and lack of comprehensive consideration about node features and external patterns. In this work, we propose three approaches to solve those three problems respectively. First, we suggest conducting similarity measure via cosine distance to take both local features and external patterns into consideration. Then, we combine the similarity measure module and the idea of adjacency-wise normalization with node-wise and batch-wise normalization and then propound partial neighborhood normalization methods to overcome the training difficulty while mitigating the impact of too much noise caused by high-density of graph. Finally, we put forward intermediate information supplement to solve the information limitation. Experiments are conducted on Yelp and Amazon datasets. And the results show that our proposed methods effectively solve the three problems. After applying the three methods, we achieve 4.81\%, 6.62\% and 6.81\% improvements in the metrics of recall, AUC and Macro-F1 respectively on the Yelp dataset. And we obtain 1.65\% and 0.29\% improvements in recall and AUC respectively on the Amazon datasets.
\end{abstract}

\keywords{Graph Neural Networks \and Fraud Detection \and Normalization}

\section{Introduction}\label{sec1}

Graph neural network (GNN) has aroused great interest in academia due to its capability that is able to handle graph-structure data and fills the vacancy in conventional deep learning field. GNN has become a significant branch of deep learning and been widely used in various fields, e.g., molecules \cite{duvenaud2015convolutional}, biological \cite{fout2017protein}, and recommend systems \cite{ying2018graph}. Many well-developed deep learning methods, for example, convolutional neural networks (CNNs) \cite{krizhevsky2012imagenet,redmon2016you} and recurrent neural networks (RNNs) \cite{ma2019ts,kenton2019bert}, which are designed to process the data with regular structure, cannot be applied directly on graph-structure data. However, the most common form of data, graph-structure data, has incalculable value to our modern society. And the demand of discovering such value has contributed to the vigorous development of GNNs. \par
GNNs can be divided into five main types \cite{zhou2020graph}. In reference \cite{zeng2021rlc}, the authors proposed an improved deep architecture for spatial-based GNNs with an application to fraud detection. The proposed method extends Camouflage-Resistant GNN (CARE-GNN) \cite{dou2020enhancing} to deep model and can divide problem into sub-problems and correct mistakes layer by layer, which forms an incremental learning mechanism. They combine layered-structure \cite{bandinelli2010learning} and residual connection \cite{he2016deep} to achieve above functions. Therefore it is named as Residual Layered CARE-GNN (RLC-GNN). In the previous work, for a given center node in a certain layer, the authors select similar 1-hop neighboring nodes by using similarity measure mechanism and aggregate the neighbors' information to the center node. In the next layer, the authors utilize the updated node representation to select a new set of neighbors. The similarity between selected nodes and the center node is higher than previous layers', which presents the self-correcting mechanism, and therefore each layer is able to focus on solving part of the problem. Since fraud detection task is essentially a node classification task and similar entities always tend to be close (i.e., nodes are directly connected by edges), they only extract information of neighboring nodes within 1-hop. \par
However, we observe that decline of the overall loss of each layer slows down starting from the fourth layer. We argue that this is because only 1-hop neighboring nodes are included during neighborhood selection and the model has made maximum use of existing information. The model cannot learn more knowledge unless new information is introduced. Besides, in the previous work, they perform similarity measure by computing the $l_1$-distance of class scores between two nodes, which has great limitations on the accuracy of similarity measure. And as we have mentioned above, RLC-GNN extends a single-layer model to multi-layer model to benefit from the self-correction mechanism and powerful representation ability of deep model. But the model also suffers from  training difficulty of deep models. \par
In this paper, we aim to provide effective approaches to address the mentioned issues, which combines: 1) We improve performance of RLC-GNN by changing the way of similarity measure from computing $l_1$-distance to  cosine distance, and we give empirical analyses of why it works; 2) For conventional deep learning methods, normalization techniques are widely used to tackle training difficulty faced by deep models. Some works \cite{li2021training, chen2020learning, zhou2020effective} has applied these useful techniques or their variants on GNNs to help training process. Inspired by these works, we propose partial neighborhood normalization methods which normalize center nodes from batch-wise or node-wise. 3) We propose an intermediate information supplement strategy to solve the problem that the use of information reaches limitation. We conduct experiments on Yelp dataset and Amazon dataset and our approaches lead to considerable performance improvements. \par

\section{Related Work}\label{sec2}

All variants of GNNs are composed of two main components which are aggregator and updater. Aggregator performs propagation operation which is to aggregate information of neighboring nodes. And updater performs transformation operation which is mainly to perform non-linear transformation to extract higher-order features. For instance, the two processes for widely used Graph Convolutional Network (GCN) \cite{Kipf:2016tc} which is a spectral-based GNN can be described as:
\begin{align}
    \rm{Aggregator: } \, N &= {\tilde D}^{ - \frac{1}{2}} \tilde A{\tilde D}^{ - \frac{1}{2}}{H^{(l - 1)}} \nonumber \\
    \rm{Updater: } \, H^{(l)} &= \sigma (N{W^{(l)}}) \label{eq1}
\end{align}
where ${\tilde A = A + {I_N}}$ is the adjacency matrix of an undirected graph ${\cal G}$ with added self-connections; $I_N$ is identity matrix; ${\tilde D_{ii}} = {\Sigma _j}{\tilde A_{ij}}$ and ${W^{(l)}}$ are layer-specific trainable weight matrices; ${H^{(l - 1)}}$ and ${H^{(l)}}$ denote the input of current layer and output respectively; $\sigma (\cdot)$ denotes an activation function. One other typical method is GraphSAGE \cite{hamilton2017inductive} which is a spatial-based GNN and its process can be formulated as:
\begin{align}
    {\rm{Aggregator: }} \, h_{{N_u}}^{(l)} &= AGGREGAT{E^{(l)}}(\{ h_v^{(l - 1)}:\forall v \in {N_u}\} ) \nonumber \\
    {\rm{Updater: }} \, h_u^{(l)} &= \sigma ({W^{(l)}}[h_u^{(l - 1)}\parallel h_{{N_u}}^{(l)}]) \label{eq2}
\end{align}
where $AGGREGAT{E^{(l)}}$ is aggregation mechanism that generates features $h_{{N_u}}^{(l)}$ by aggregating features of a set of neighboring nodes ${N_u}$ to center node $v$ at current layer; $h_v^{(l - 1)}$ and $h_u^{(l - 1)}$ denotes input feature at current layer of a neighboring node and the center node respectively; $h_u^{(l)}$ denotes output feature of the center node at current layer; $[\cdot \parallel \cdot]$ is vector concatenating operation. The spatial-based GNNs, represented by GraphSAGE, have great flexibility in the design of aggregation mechanism, which will greatly benefit the performance of downstream tasks. Therefore, RLC-GNN are designed with spatial-based architecture. In this work, we further improve the performance of RLC-GNN by optimizing neighboring nodes selector which is part of aggregator. \par
Recent work \cite{liu2020towards} reveals that although over-smoothing issue contributes to the performance deterioration to deep GNNs, the entanglement of propagation and transformation is the key factor. And Deep Adaptive Graph Neural Network (DAGNN) is proposed to adaptively incorporate information from different size of receptive fields on graphs by decoupling propagation and transformation and applying attention mechanism. And this inspires us to propose the intermediate information supplement to solve the insufficient information problem which blocks the learning process of RLC-GNN. \par
Normalization techniques are critical methods that are widely used in conventional deep learning field to overcome training difficulty of deep neural networks, e.g., Batch Normalization (BatchNorm) \cite{ioffe2015batch} and Layer Normalization (LayerNorm) \cite{ba2016layer}. BatchNorm first computes batch statistics including mean and variance along batch dimension and then normalizes each scalar feature independently. LayerNorm is a variant of BatchNorm. And the difference is that LayerNorm computes statistical information along channel dimension. Some works \cite{chen2020learning, zhou2020effective} introduce existing normalization methods or propose novel ones to train deep GNNs. Work \cite{chen2020learning} investigates existing graph normalization methods and formulates them into four levels: node-wise, adjacency-wise, graph-wise, and batch-wise. Inspired by this work, we combine our similarity measure mechanism and the idea of adjacency-wise normalization with two methods, batch-wise normalization and node-wise. And we propose partial neighborhood normalization on batch-wise and node-wise. \par
Performing fraud detection on graph-structure data is essentially solving a node classification problem. However, fraud detection task has some peculiarities. First, there are far less fraudsters in samples, which makes it hard for models to learn features of fraudsters drown in lots of benign samples. Additionally, smart fraudsters always evade detectors by actively camouflaging their features (e.g., replacing keywords with special symbols but semantics remaining the same) \cite{dou2020enhancing}, which makes it even more difficult for models to recognize features of fraudsters. CARE-GNN performs label-aware similarity measure to confront camouflages. More precisely, it computes $l_1$-distance between a given center nodes and neighboring nodes. However, due to the complex behavior of fraudsters, $l_1$-distance is not accurate enough to describe the characteristics of fraudsters, which limits the recognition ability of models. We propose to use cosine-distance instead of $l_1$-distance to perform similarity measure between a pair of nodes. \par

\section{Methodology}\label{sec3}

In this section, we introduce three methods to solve three issues of RLC-GNN, respectively. Based on the 6-layers RLC-GNN, we improve the original model for fraud detection task. Given a heterogeneous graph ${\cal G} = \{ \mathcal{V}, \mathcal{X},\{ {\mathcal{E}_r}\} \vert_{r = 1}^R\}$, $\mathcal{V}$ is the set of nodes' indices. Each node $v_i$ is represented by a $d$-dimension feature vector ${x_i} \in {\mathbb{R}^d}$ in set $\mathcal{X}$ . Each edge $e_{ij}^r \in {\mathcal{E}_r}$ indicates node $v_i$ and $v_j$ are connected by an edge under relation $r \in \{ 1,2, \cdots ,R\}$.

\subsection{Similarity Measure via Cosine Distance}\label{subsec3-1}

We apply cosine distance instead of $l_1$-distance to perform similarity measure between two given nodes. And we give intuitive analysis for this change. Yelp dataset used in this paper consists of reviews submitted by users on Yelp websites. Given a certain review under a product, the features include its ranking among all reviews under the product, tendency (very high rating or low), the time when it is submitted, uniqueness (whether it is the only review given by the user under the product), length, ratio of competent vocabularies to objective vocabularies, etc. Fraud detection is a semi-supervised node classification task, which means that we let our model to learn the representation of fraudulent samples (i.e., what fraudulent samples themselves look like and what types of topologies they will form with their neighbors) and then to find similar samples based on learned patterns. Intuitively, each person has his own style, even the fraudsters. And the reviews submitted by the same fraudster should have similar features. A fraudster may deliberately change his language style of his reviews, which makes model unable to find common semantic features. But except these local differences, the overall external characteristics (i.e., subconscious behaviors) of fraudsters stay the same, which results in their reviews having similar topological structure on the built graphs. We need to evaluate the overall indication of samples rather than numerical difference between feature vectors. The model will benefit from more comprehensive consideration of features and capturing of similar structure, and a fully connected layer with $tanh$ activation function are applied to extract topological structure features in this work. \par
In mathematics, cosine similarity measures the difference between two vectors by computing the cosine value of the angle between the two vectors. The closer the angle is to zero (i.e., cosine value is close to one), the closer the two vectors are. Cosine distance is usually used to analyze user behavior in practical applications, e.g., recommend systems. In this work, a review or a user is represented by an embedding vector. And we need to measure the similarity between samples from the perspective of user behavior. Thus, we optimize the origin method to use cosine distance instead. For a center node $u$ and any one of its neighboring node $v$' at $l$-th layer, the cosine similarity is formulated as:
\begin{equation}
    S^{(l)}{(u,v^{'})} = \frac{\sigma (ml{p^{(l)}}(h_u^{(l - 1)})) \cdot \sigma (ml{p^{(l)}}(h_{v'}^{(l - 1)}))}{{{\left\| {\sigma (ml{p^{(l)}}(h_u^{(l - 1)}))} \right\|}_2}{{\left\| {\sigma (ml{p^{(l)}}(h_{{v'}}^{(l - 1)}))} \right\|}_2}} \label{cosd}
\end{equation}
where $h_u^{(l - 1)}$ and $h_{v'}^{(l - 1)}$ denote the input feature vectors of a center node and a neighboring node respectively. And then we can define the cosine distance between node $v$ and $u$ as:
\begin{equation}
    {\mathcal{D}^{(l)}}(u,{v'}) = 1 - {S^{(l)}}(u,{v'})    \label{cossim}
\end{equation}
Each layer has its own similarity measure module.

\subsection{Partial Neighborhood Normalization}\label{subsec3-2}
Original normalization techniques on graph compute statistics without any filtering mechanism. For instance, batch normalization computes mean and variance among batch dimension after the aggregation of all neighbors and node normalization among node dimension. Of particular note is that fraud detection is a node classification task with the problem of severe unbalance in quantitative terms. With that in mind, we consider combine the existing similarity measure and the idea of adjacency normalization with BatchNorm and NodeNorm to avoid feature distortion which is caused by adding too much benign samples' information when performing normalization on fraudsters' features. Because BatchNorm computes along batch dimension, more noise will be introduced, especially on dense graph. The distortion will be amplified, and model will suffer performance deterioration instead of benefitting from applying batch-wise normalization. \par
By combining similarity measure mechanism with normalization, for fraudulent samples, a percentage of dissimilar samples will be discarded before performing normalization, which alleviates the distortion. For a center node $u$ and all its neighboring nodes $N(u)$ at $l$-th layer, we define the node-wise partial neighborhood normalization on heterogenous graph as follows:
\begin{align}
    {\mu _u} &= \frac{1}{{d \cdot \vert \mathbb{N}(u)\vert }}\sum\limits_{i = 1}^d {\sum\limits_{r = 1}^R {p_r^{(l)}\sum\limits_{v \in \mathbb{N}(u)} {h_{v,r,i}^{(l)}} } }  \nonumber \\ 
    \sigma _u^2 &= \frac{1}{{d \cdot \vert \mathbb{N}(u)\vert }}\sum\limits_{i = 1}^d {[{(\sum\limits_{r = 1}^R {p_r^{(l)}\sum\limits_{v \in \mathbb{N}(u)} {h_{v,r,i}^{(l)}) - {\mu _u}} } ]^2}} \nonumber \\ 
    \hat h_u^{(l)} &= \frac{{h_u^{(l)} - broadcast({\mu _u})}}{{\sqrt {\sigma _u^2 + \varepsilon } }} \label{nodewise}
\end{align}
where ${\mu _u}$ and ${\sigma _u}$ are the mean and variance computed over features of center node $u$ itself; $h_{v,r}^{(l)} = [h_{v,r,1}^{(l)}, \cdots ,h_{v,r,d}^{(l)}]$ is the $d$-dimension node feature vector of a neighboring node $v$ at $l$-th layer; $p_r^{(l)}$ is the threshold for relation $r$; $\mathbb{N}(u) \subseteq N(u)$ denotes neighboring nodes being selected after similarity measure and $\vert \mathbb{N}(u)\vert $ is the number of neighbors; $\hat h_u^{(l)}$ is the normalized feature of center node; $broadcast({\mu _u})$ is broadcast operation which repeats the mean value to form a vector with the same shape as $h_u^{(l)}$. Similarly, the definition of batch-wise partial neighborhood normalization is shown in Eq.\ref{batchwise}:
\begin{align}
    {\mu _{B,i}} &= \frac{1}{m}\sum\limits_{j = 1}^m {\frac{1}{{\vert \mathbb{N}({u_j})\vert }}\sum\limits_{r = 1}^R {p_r^{(l)}\sum\limits_{v \in \mathbb{N}({u_j})} {h_{v,r,i,{u_j}}^{(l)}} } } \nonumber \\ 
    \sigma _{B,i}^2 &= \frac{1}{m}\sum\limits_{j = 1}^m {\frac{1}{{\vert \mathbb{N}({u_j})\vert }}[{(\sum\limits_{r = 1}^R {p_r^{(l)}\sum\limits_{v \in \mathbb{N}({u_j})} {h_{v,r,i,{u_j}}^{(l)}) - {\mu _{B,i}}} } ]^2}} \nonumber \\ 
    \hat h_{{u_j},i}^{(l)} &= \frac{{h_{{u_j},i}^{(l)} - {\mu _{B,i}}}}{{\sqrt {\sigma _{B,i}^2 + \varepsilon } }} \label{batchwise}
\end{align}
where ${\mu _{B,i}}$ and ${\sigma _{B,i}}$ are the mean and variance of the $i$-th feature computed over batch dimension; $h_{v,r,i,{u_j}}^{(l)}$ denotes the neighboring node's $i$-th feature of the $j$-th center node in the batch under relation $r$; $m$ is the size of batch; $\hat h_{{u_j},i}^{(l)}$ is the normalized $i$-th feature of the $j$-th center node. It can be noted that, compared with conventional BatchNorm, there is no scaling and shifting operations.

\subsection{Intermediate Information Supplement}\label{subsec3-3}

Given a center node painted red, we show the processes in a layer with 6-layers RLC-GNN in Figure 2\ref{sub@subfig2-1}. One color denotes that a neighboring node is connected to the center node under a relationship corresponding to the color. The heterogeneous graph is converted to three homogeneous sub-graphs. Then we compute the $l_1$-distances between center node and all neighbors under each relation respectively and sort the distances in ascending order. All distances will be fed through a reinforcement learning module to update the thresholds (i.e., preservation ratio of nodes under each relation. See \cite{dou2020enhancing} for more details). And then the nodes which are not similar enough will be discarded. Now, all nodes to be used for aggregation under each relation is determined. \par
\begin{figure}
    \centering
    \includegraphics[width=0.55\textwidth]{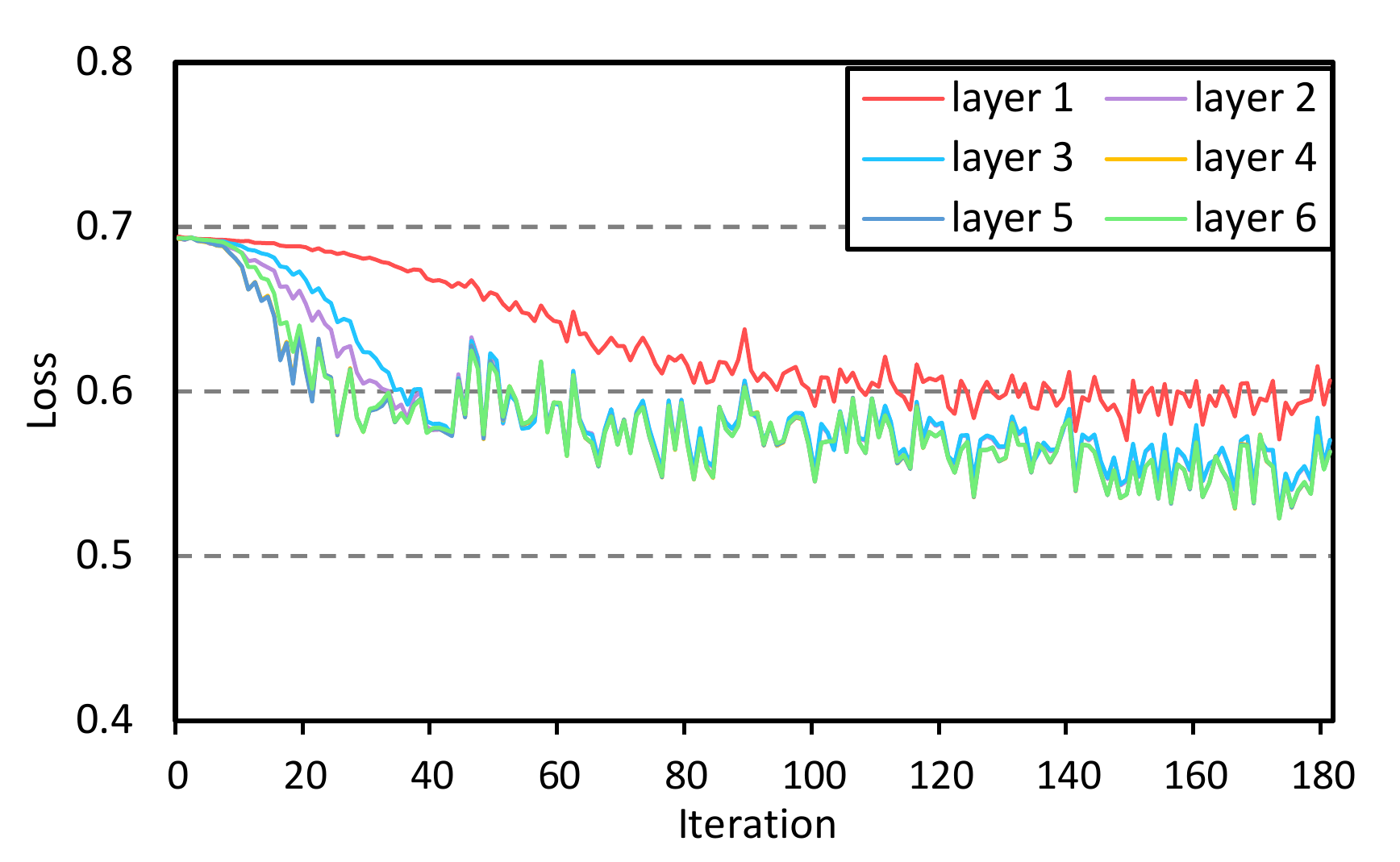}
    \caption{Training loss of each layer on Yelp dataset with 6-layers RLC-GNN.}\label{fig1}
\end{figure}

The training loss of each layer with 6-layers RLC-GNN decreases notably in the first four layers. However, the descend of overall layer training loss after the fourth layer becomes small (shown in Figure \ref{fig1}). We argue that it is caused by the limited information of 1-hop neighboring nodes. The first four layers have made the maximum use of information. With the incremental learning mechanism, a layer inherits the knowledge learned by its previous layers from graph. Empirically, new information is required for further learning. Otherwise, what the last two layers can do under the original design is simply delivering the knowledge learned by previous layers. To alleviate this issue, we expand the range for extracting information to 2-hop starting from the fourth layer to introduce new information for further learning, for the reason of which we name it as intermediate information supplement (IIS). As we have mentioned before, fraud detection is a node classification task, and similar nodes tend to be close (relationship rather than spatial distance). Therefore, we do not need to consider nodes far away from center node, which is markedly different from graph-level classification tasks that global information needs to be considered. An illustration of intermediate information supplement is provided in Figure 2\ref{sub@subfig2-2}. \par

\begin{figure}
    \centering
    \begin{minipage}{1.0\linewidth}
        \subfigure[Overview of layer structure after applying IIS in RLC-GNN. The IIS is added before intra-relation aggregation module only after the third layer. In the original model, all layers only aggregate features from 1-hop neighboring nodes. IIS enables deep layers to extract richer information for further learning.]{
            \includegraphics[width=1.0\textwidth]{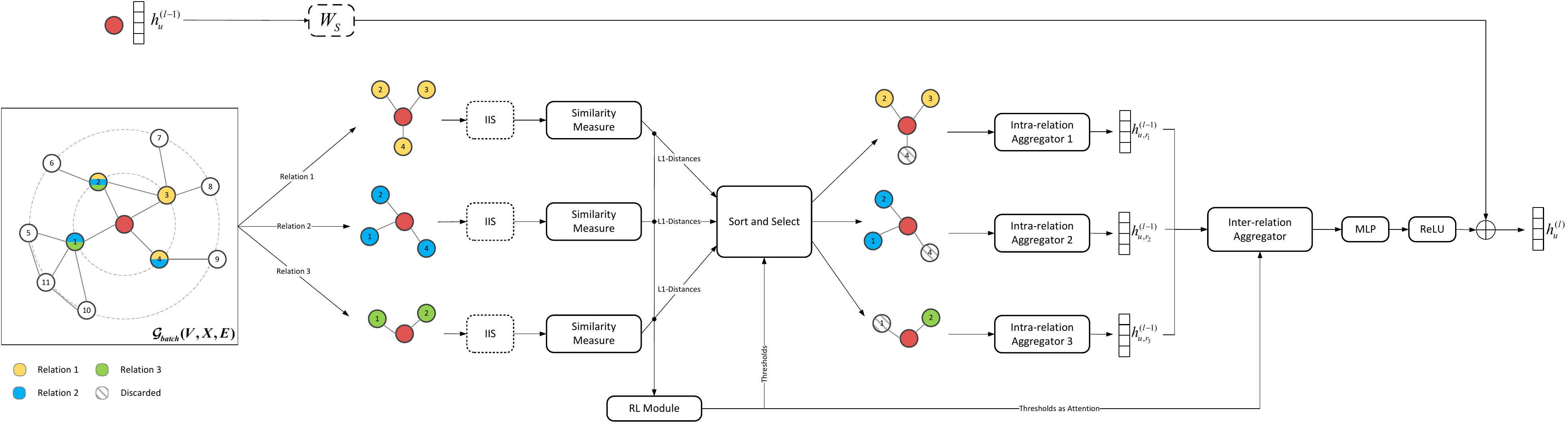}
            \label{subfig2-1}
        }
    \end{minipage}
    \vfill
    \begin{minipage}{1.0\linewidth}
        \subfigure[Process of IIS. We take the IIS process of the given center node under relation 1 as example to illustrate how IIS works. In all layers, we get all 1-hop neighboring nodes represented by set $A=\{ v, v\in N{(u)}\}$, where $u$ is the center node and $N{(\cdot)}$ denotes the set of 1-hop neighbors of $u$. Then starting from the fourth layer, we make all nodes in $A$ take turns to be center node temporarily to find neighboring nodes under relation 1 within 1-hop (i.e., we expand the receipt field of center node $u$ to 2-hop, which more information is introduced) and we get an expanded sub-graph under relation 1.]{
            \includegraphics[width=1.0\textwidth]{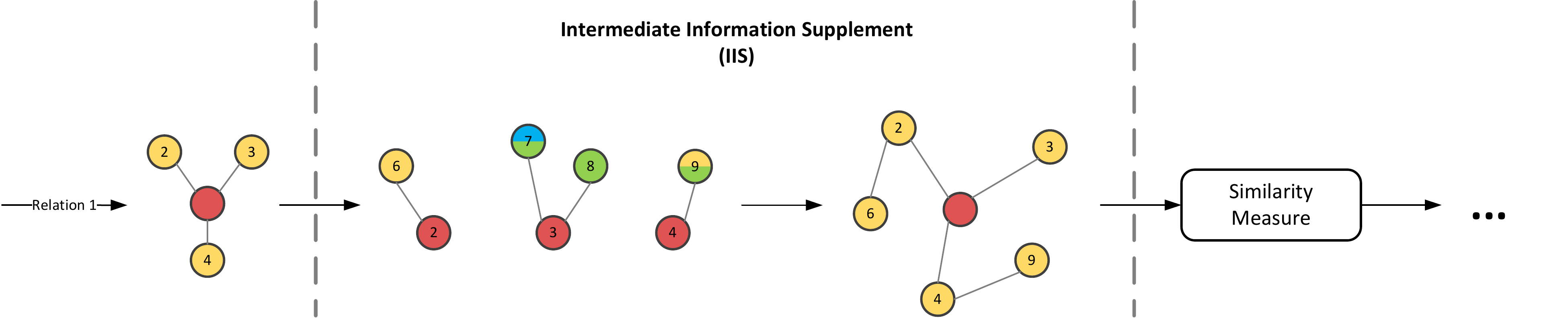}
            \label{subfig2-2}
        }
    \end{minipage}
    \caption{Our proposed intermediate information supplement (IIS). In Figure 2\subref{subfig2-1}, we show the overview of layer structure of RLC-GNN with IIS. And we illustrate the IIS process by taking the example under a relation in Figure 2\subref{subfig2-2}. IIS is an idea that it is not only applicable to RLC-GNN, but also empirically to any GNN that performs information aggregation locally.}
    \label{fig2}
\end{figure}

\section{Experiments}\label{sec4}

\subsection{Dataset}\label{subsec4-1}

We conduct experiments on Yelp dataset including hotel and restaurant reviews and Amazon dataset including users under Musical Instruments category, which is the same as that in work \cite{zeng2021rlc}. For Yelp dataset, it consists of 45,954 reviews, of which 14.5\% are fraudulent samples. And Amazon dataset contains 11,944 users, of which 9.5\% are fraudsters. Each review is a node represented by a vector composed of 32 handcraft features for Yelp (25 for Amazon). Three relations are designed for each dataset (i.e., there are three types of edges). If two samples satisfy a relationship, they are connected by a type of edge corresponding to the relation. Yelp: (1) R-U-R: two reviews are submitted by a same user; (2) R-S-R: two reviews give the same star rating under a product; (3) R-T-R: two reviews are submitted in a same month under a product. Amazon: (1) U-P-U: Two users rate at least one same product; (2) U-S-U: Two users give at least one same star rating under a product in a week; (3) U-V-U: The text similarity of mutual reviews between two users reaches the 5\% among all users. In this design, Yelp dataset has 3,846,979 edges in total and Amazon dataset 4,398,392.

\subsection{Implementation and Setup}\label{subsec4-2}

All methods are implemented with Pytorch 1.7.0. Batch size is set to 256 for Amazon and 512 for Yelp. The training set accounts for 40\% on both datasets. And the optimization method is Adam \cite{kingma2015adam}. All experiments are running on Python 3.7.6, AMD Ryzen7 4800H CPU and a single RTX 2060 GPU. \par
We conduct experiments based on RLC-GNN-6. Three methods take turns to be added to the model to verify the effectiveness of each method, and we conduct performance comparison of the model combined with all three proposed methods to various GNNs. To begin with, we verify effectiveness of performing similarity measure via cosine distance. Afterwards we show the results of partial neighborhood normalization methods among node-wise and batch-wise. In the next part, we put intermediate information supplement to the test based on the RLC-GNN with cosine distance measuring module and normalization technique. Finally, we give the performance of RLC-GNN with all methods proposed in this work. \par
With regard to selection of evaluation metrics, we use recall, macro-F1 and AUC \cite{powers2020evaluation} to assess our proposed methods. Recall directly assess the ability of recognizing fraudulent samples. Macro-F1 is picked up to avoid the misleading which our model achieves high recall by simply predicting all samples as frauds. AUC is chosen for its insensitivity on sample distribution. \par

\subsection{Result}\label{subsec4-3}

\subsubsection{Similarity Measure via Cosine Distance}\label{sssec4-3-1}

Given two feature vectors, we first use layer-specific MLP to transform the dimension of features to 8 and then we compute cosine distance (Eq.\ref{cosd}). In Table \ref{tab1}, we show the experiment results of RLC-GNN with similarity measuring via computing cosine distance. We achieve improvements in almost all metrics on both datasets. But macro-F1 on Amazon dataset shows that the performance of our model has been sacrificed somewhere. Macro-F1 is the harmonic average of recall and precision which denotes the percentage of correct predictions on samples whose predictions are positive. The fact that we get higher recall, meanwhile getting lower marco-F1, indicates that the precision of the model decreases. The higher recall suggests that the model can recognize more fraudulent samples by judging samples more comprehensively, which empirically validates the discussion in Section \ref{subsec3-1}. The lower precision denotes that the model becomes more aggressive which we argue that it is relevant to some statistics of the graph and will be discussed in next part.
\begin{table}[h]
    \begin{center}
    \begin{minipage}{\textwidth}
    \caption{Results of RLC-GNN-6 with $l_1$-distance and cosine distance on both datasets. We report best results after training 200 epochs. The \emph{reference} model is RLC-GNN-6 with similarity measure via $l_1$-distance.}\label{tab1}
    \begin{tabular*}{\textwidth}{@{\extracolsep{\fill}}lcccc@{\extracolsep{\fill}}}
        \toprule
        \textbf{Dataset} & \textbf{Model} & \textbf{Recall(\%)} & \textbf{AUC(\%)} & \textbf{Macro-F1(\%)} \\
        \midrule
        \multirow{2}{*}{Yelp}
        & \emph{Ref.}   & 74.66  & 83.29 & 68.45 \\
        & w/ cos distance   & \textbf{75.68}  & \textbf{84.19} & \textbf{68.81} \\
        \midrule
        \multirow{2}{*}{Amazon}
        & \emph{Ref.}   & 89.83  & 96.77 & \textbf{90.08} \\
        & w/ cos distance   & \textbf{90.62}  & \textbf{97.19} & 89.21 \\
        \bottomrule
    \end{tabular*}
    \end{minipage}
    \end{center}
\end{table}

\subsubsection{Partial Neighborhood Normalization}\label{sssec4-3-2}

\begin{figure}
    \centering
    \begin{minipage}{0.49\linewidth}
        \subfigure[Training loss on Yelp dataset.]{
            \includegraphics[width=1.0\linewidth]{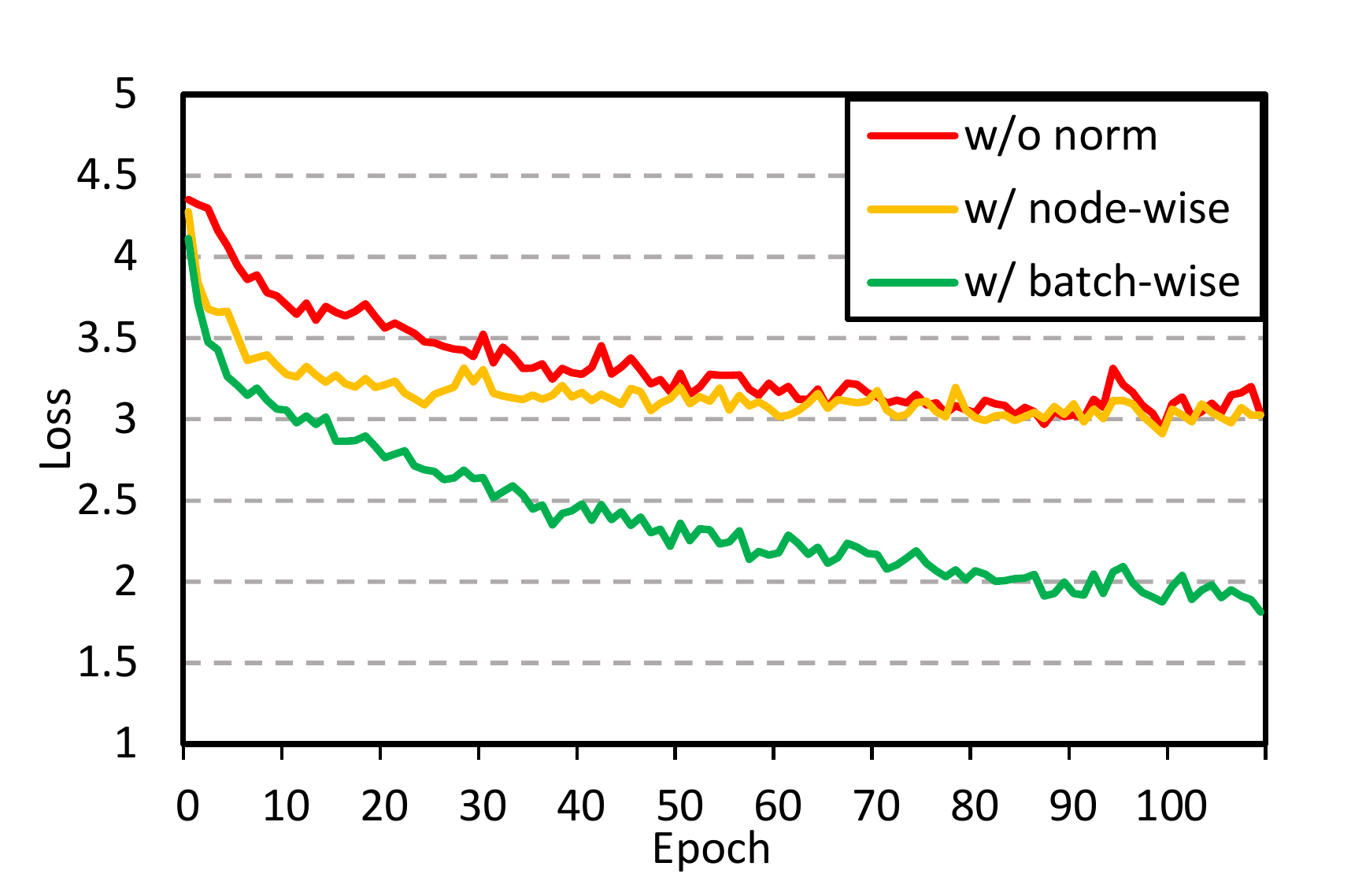}
            \label{subfig3-1}
        }
    \end{minipage}
    \begin{minipage}{0.49\linewidth}
        \subfigure[Traing loss on Amazon dataset.]{
            \includegraphics[width=1.0\linewidth]{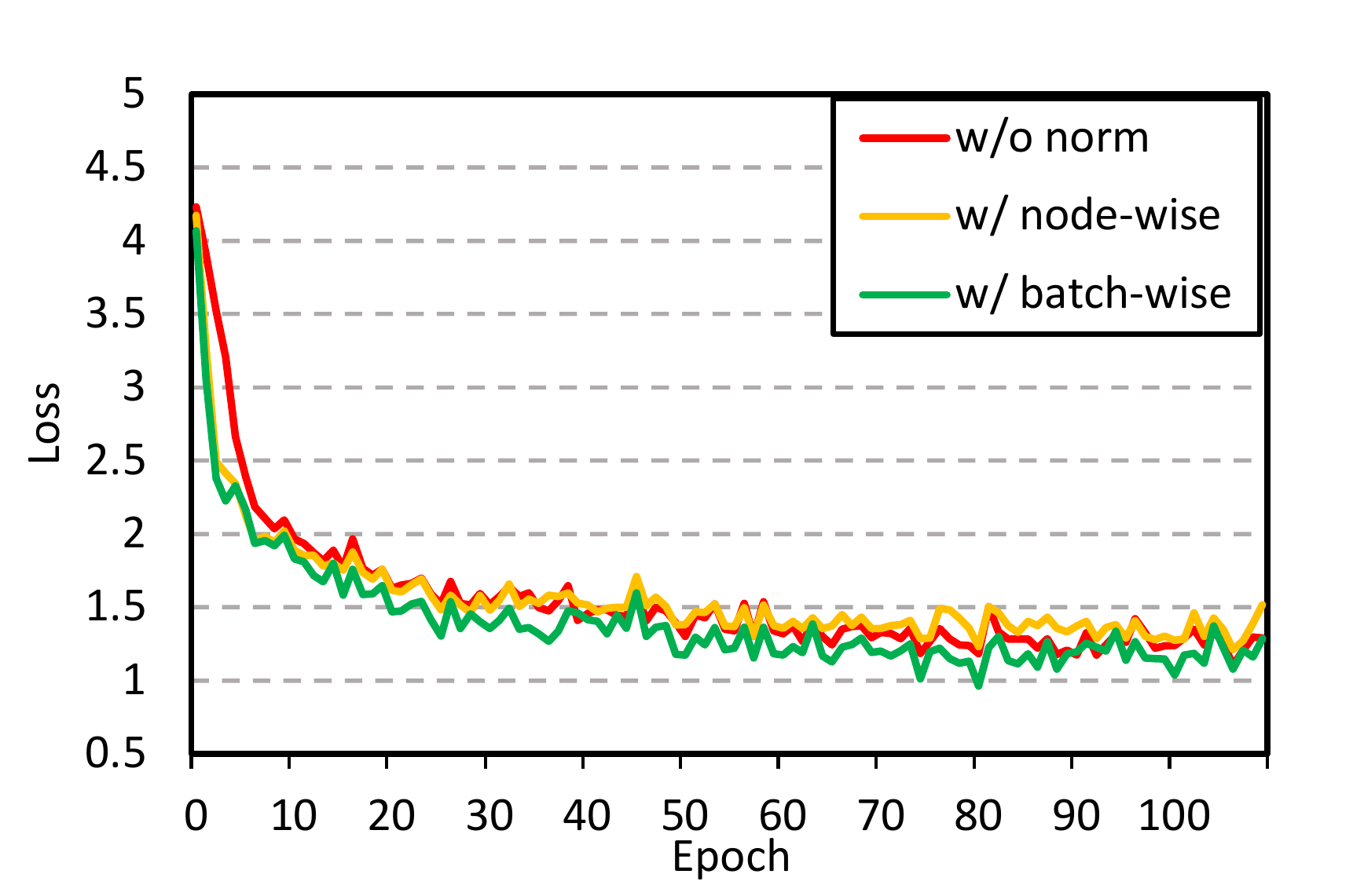}
            \label{subfig3-2}
        }
    \end{minipage}
    \caption{Training loss of RLC-GNN without normalization and with node-wise normalization or batch-wise. Obviously, both node-wise and batch-wise normalization methods can accelerate training process. And batch-wise normalization can even help optimizer reach much better solutions.}
    \label{fig3}
\end{figure}

For a given center node, we first perform similarity measure and then select similar neighboring nodes. Next, we aggregate neighboring features to the center node when we have completed all preparations. Finally, we conduct normalization along node-wise or batch-wise. We show the training loss in Figure \ref{fig3}. We notice that normalization methods accelerate the training process, especially on Yelp dataset. For node-wise partial neighborhood normalization, it can make the model converge faster, but not better. And for batch-wise normalization, it accelerates training process even faster, and better weights are found in solution space. However, on Amazon dataset, applying normalization along batch-wise causes some damage to the model which can be seen from macro-F1. We show the experiments results in Table \ref{tab2}. The interesting note is that we significantly accelerate the training with normalization along both wises on Yelp dataset, while the fraud detection task on Amazon dataset suffers applying batch-wise normalization. More specifically, the macro-F1 has significant decline. And because recall and AUC stay roughly the same, which means that the ability of recognizing frauds remains basically the same, we can know that the precision has significant decline according to the computation of macro-F1. We investigate this phenomenon from the perspective of statistics of the graphs and we provide empirical analysis as follows. \par
\begin{figure}
    \centering
    \includegraphics[width=0.57\textwidth]{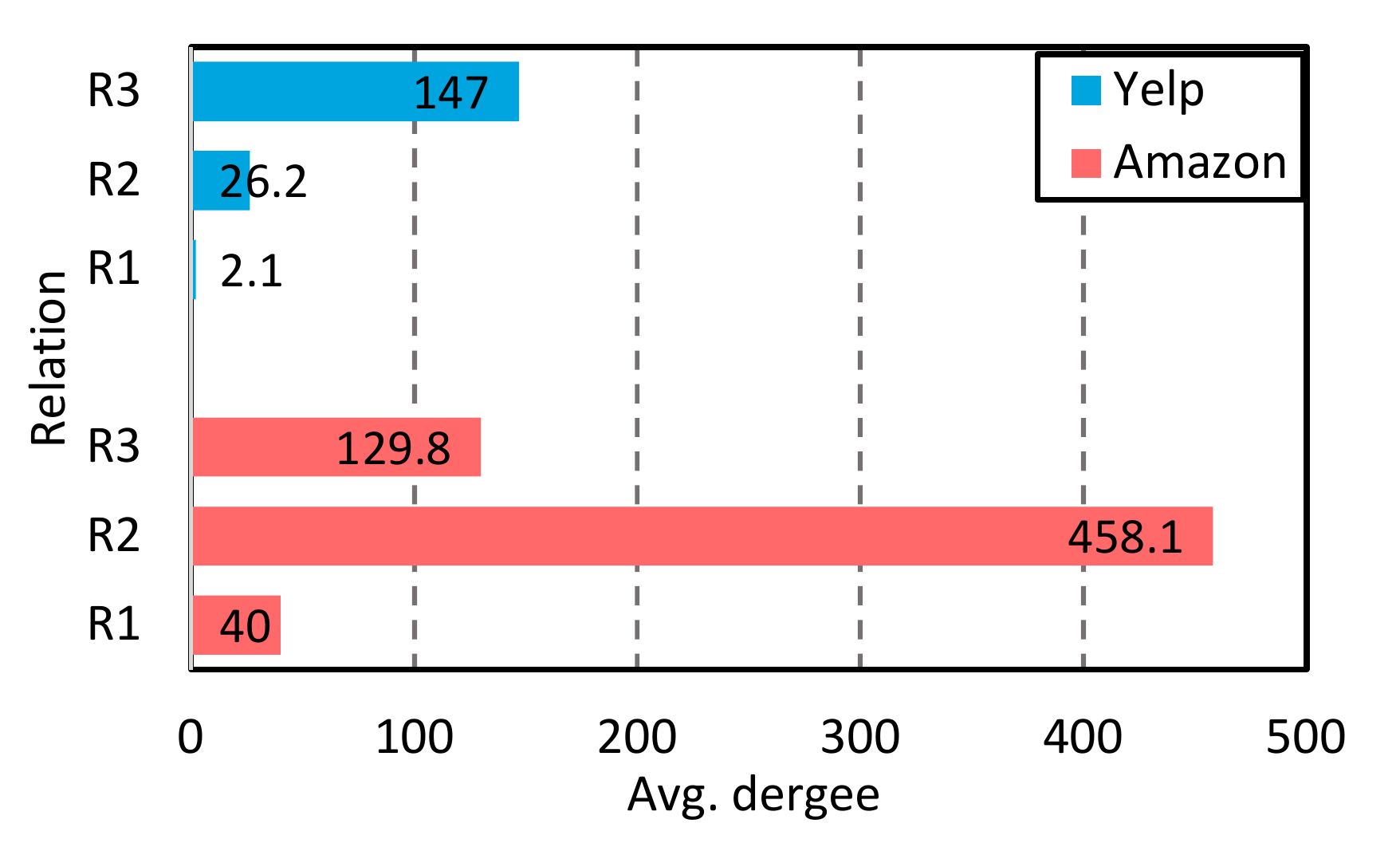}
    \caption{The average degree of nodes under each relation in Yelp and Amazon datasets. It is noticeable that the graph built upon Amazon dataset is much denser than Yelp.}
    \label{fig4}
\end{figure}

In Figure \ref{fig4}, we show the average degree of nodes under each relation in both datasets. Except the average degrees under the third relation are close, the degrees under other relations in Amazon is nearly 19 times greater than Yelp, which means graph built upon Amazon is much denser than Yelp. Due to the denseness of graph and high proportion of benign samples, the graph is full of information of benign features and too many benign samples are gathered around fraudulent samples, which leads to the fact that features of fraudulent samples include too many benign samples' information after aggregation. In Section \ref{subsec3-1}, we propose to apply cosine distance to perform similarity measure to capture patterns from the perspective of overall behavior (i.e., the topological structure in graph). However, because of the high density, the topological structures of all nodes become similar. Driven by supervised learning mechanism, the model not only learns normally, but also learns treating benign samples as fraudulent samples. The normalization techniques do not improve models' ability, but only allows models to acquire the ability they should have earlier. We indicate in Section \ref{subsec3-2} that too much noise (i.e., information of benign samples) will amplify the distortion of original features. For the batch-wise normalization, more unrepresentative information is included in the computation, which makes fraudulent samples be more like benign samples. In summary, these reasons make the model perform more radical under the interaction. For the node classification task, in special in category imbalance, we suggest taking the density of graph into consideration when deciding what type of normalization technique to be used. Based on the results in this work, we recommend batch-wise normalization for sparse graph and node-wise for dense. \par

\begin{table}
    \begin{center}
    \begin{minipage}{\textwidth}
    \caption{Results based on the RLC-GNN with similarity measure via cosine distance which is the \emph{reference} model. We report the best results after training 200 epochs.}\label{tab2}
    \begin{tabular*}{\textwidth}{@{\extracolsep{\fill}}lcccc@{\extracolsep{\fill}}}
        \toprule
        \textbf{Dataset} & \textbf{Model} & \textbf{Recall(\%)} & \textbf{AUC(\%)} & \textbf{Macro-F1(\%)} \\
        \midrule
        \multirow{2}{*}{Yelp}
        & \emph{Ref.}   & 75.68  & 84.19  & 68.81 \\
        & w/ node-wise norm  & 76.71  & 85.07 & 68.87 \\
        & w/ batch-wise norm  & \textbf{78.03}  & \textbf{88.53}  & \textbf{74.72} \\
        \midrule
        \multirow{2}{*}{Amazon}
        & \emph{Ref.}   & 90.62  & \textbf{97.19} & 89.21 \\
        & w/ node-wise norm  & \textbf{91.48}  & 97.06 & \textbf{89.42} \\
        & w/ batch-wise norm  & 89.70  & 96.53 & 76.47 \\
        \bottomrule
    \end{tabular*}
    \end{minipage}
    \end{center}
\end{table}

\subsubsection{Intermediate Information Supplement}\label{sssec4-3-3}

Based on the discussion in Section \ref{subsec3-3}, we expand the receipt field to 2-hop starting from the fourth layer. In view of the analysis in Section \ref{sssec4-3-2}, we take the model with cosine similarity measure module and batch-wise partial neighborhood normalization as reference model on Yelp dataset. As for the Amazon dataset, due to its high density, we do not further import more information, which too much benign samples lack of crucial information will actually become noise. We show the experiment results in Table \ref{tab3}. The model applying intermediate information supplement outperforms the reference model in all metrics.
\begin{table}[h]
    \begin{center}
    \begin{minipage}{\textwidth}
    \caption{Experiment results of the model with intermediate information supplement and batch-wise normalization based on the \emph{reference} model on Yelp dataset. We report the best results after training 200 epochs. IIS is abbreviation of intermediate information supplement.}\label{tab3}
    \begin{tabular*}{\textwidth}{@{\extracolsep{\fill}}lcccc@{\extracolsep{\fill}}}
        \toprule
        \textbf{Dataset} & \textbf{Model} & \textbf{Recall(\%)} & \textbf{AUC(\%)} & \textbf{Macro-F1(\%)} \\
        \midrule
        \multirow{2}{*}{Yelp}
        & \emph{Ref.}   & 78.03  & 88.53 & 74.72 \\
        & w/ IIS   & \textbf{79.47}  & \textbf{89.91} & \textbf{75.26} \\
        \bottomrule
    \end{tabular*}
    \end{minipage}
    \end{center}
\end{table}

\subsubsection{Comparison}\label{sssec4-3-4}

In this section, we conduct performance comparison of the model applying all methods proposed in this work to various GNNs on the two datasets. We show the experiments results in Table \ref{tab4}. Note that, on Yelp dataset, the model with all proposed methods significantly outperforms the basic RLC-GNN-6, and it even outperforms 27-layers model a lot, which validates the effectiveness of our methods. As for Amazon dataset, applying proposed methods makes the model outperform the basic 6-layers model in recall and AUC. We have analyzed the causes of the decline of macro-F1 and argue that similarity measure via cosine distance makes the model become more radical on dense graph. In particular, it takes about 13.9 milliseconds for RLC-GNN-6 with cosine similarity measuring module and node-wise normalization to conduct inference for a node and 53.4 milliseconds for RLC-GNN-27. In other words, in order to achieve the performance to that of the 27-layers RLC-GNN in the task of finding out fraudulent samples, the efficiency can be increased by about 4 times. However, if we are very concerned about the misjudgment that labels benign samples as fraudulent, the best way to improve performance is stacking more layers. It is a tradeoff.
\begin{table}
    \begin{center}
    \begin{minipage}{\textwidth}
    \caption{Performance comparison of RLC-GNN-6 with all proposed methods to various GNNs. Other models' results are directly quoted from the works \cite{zeng2021rlc,dou2020enhancing}.}\label{tab4}
    \begin{tabular*}{\textwidth}{@{\extracolsep{\fill}}lcccc@{\extracolsep{\fill}}}
        \toprule
        \textbf{Dataset} & \textbf{Model} & \textbf{Recall(\%)} & \textbf{AUC(\%)} & \textbf{Macro-F1(\%)} \\
        \midrule
        \multirow{2}{*}{Yelp}
        & GCN   & 50.81  & 54.47  & -- \\
        & GraphSAGE  & 52.86  & 54.00 & -- \\
        & GAT  & 54.52 & 56.24 & -- \\
        & CARE-GNN  & 71.02  & 77.72 & 61.13 \\
        & RLC-GNN-6  & 74.66  & 83.29 & 68.45 \\
        & RLC-GNN-27  & 76.68  & 85.44 & 70.03 \\
        & \makecell{\textbf{\emph{w/ cos distance,}} \\ \textbf{\emph{batch-wise norm \rm{\&} IIS}}}  & \textbf{79.47}  & \textbf{89.91} & \textbf{75.26} \\
        \midrule
        \multirow{2}{*}{Amazon}
        & GCN  & 67.45  & 74.34  & -- \\
        & GraphSAGE  & 70.16  & 75.27 & -- \\
        & GAT  & 65.51 & 75.16 & -- \\
        & CARE-GNN  & 88.17  & 93.21 & 87.81 \\
        & RLC-GNN-6  & 89.83  & 96.77 & \textbf{90.08} \\
        & RLC-GNN-27  & \textbf{91.83}  & \textbf{97.48} & 89.18 \\
        & \makecell{\textbf{\emph{w/ cos distance \rm{\&}}} \\ \textbf{\emph{node-wise norm}}} & 91.48  & 97.06 & 89.42 \\
        \bottomrule
    \end{tabular*}
    \end{minipage}
    \end{center}
\end{table}



\section{Conclusion}\label{sec5}

In this work, we find three issues, which are the lack of comprehensive consideration about features and neighborhood topological structure, training difficulty of deep models and the use of information reaching limitation, in the previous work which conducts fraud detection via advanced RLC-GNN. We propose three methods to solve each issue respectively. First, we propose to perform similarity measure via cosine distance to learn node features more comprehensively. Then we apply batch-wise or node-wise partial neighborhood normalization to accelerate the training of deep RLC-GNN. And according to the experiments results, we achieve significant improvements on Yelp dataset. We improve the ability of recognizing fraudulent samples. However, we find that the models become more radical after applying proposed methods on Amazon dataset. We provide an empirical analysis and argue that it is mainly caused by high-density of the graph. And we suggest taking the graph density into consideration when we try to apply normalization techniques. If the graph is very dense, it is good to choose node-wise normalization for its not introducing more invalid information, especially when there is a problem of category imbalance. Finally, we use intermediate information supplement to extract more information starting from intermediate layer to help the model learn further. And the effectiveness of all proposed methods is validated by experiments conducted on Yelp and Amazon datasets. \par

\bibliographystyle{unsrt}  
\bibliography{references}

\end{document}